\pdfoutput=1

\documentclass[11pt]{article}
\usepackage{cite}
\usepackage[]{acl}

\usepackage{times}
\usepackage{latexsym}

\usepackage[T1]{fontenc}

\usepackage[utf8]{inputenc}

\usepackage{microtype}

\usepackage{color}
\usepackage{amsmath}
\usepackage{graphicx}
\usepackage{arydshln} 
\usepackage{multirow}

\definecolor{c1}{HTML}{95bddc}
\definecolor{c2}{HTML}{c2d1e5}
\definecolor{c3}{HTML}{fe793d}
\definecolor{c4}{HTML}{fb4c1f}
\definecolor{c5}{HTML}{b71a3b}
\definecolor{c6}{HTML}{7e0f12}
\definecolor{c7}{HTML}{E85642}
\definecolor{c8}{HTML}{C00000}
\definecolor{c9}{HTML}{ff2d51}
\definecolor{blue}{HTML}{339ff4}
\definecolor{green}{HTML}{3ca057}
\title{An Iterative Associative Memory Model for \\ Empathetic Response Generation}

\author{
Zhou Yang$^{1}$, Zhaochun Ren$^{2}$, Yufeng Wang$^{1}$, \textbf{Haizhou Sun$^3$}, \\
\textbf{Chao Chen$^{4}$}, \textbf{Xiaofei Zhu$^{5}$},
\textbf{Xiangwen Liao}$^{1}$\thanks{\hspace{1mm} Corresponding author.}\\
\small $^1$College of Computer and Data Science, Fuzhou University,
 Fuzhou, China; 
\small $^2$Leiden University, Leiden, The Netherlands \\
\small $^3$H. Sun is with SmartMore;
\small $^4$ School of Computer Science and Technology, Harbin Institute of Technology, Shenzhen, China \\
\small $^5$College of Computer Science and Technology, Chongqing University of Technology, Chongqing, China\\ 
\small  \texttt{\{200310007, 211027083, 102102153, liaoxw\}@fzu.edu.cn} \\
\small  \texttt{z.ren@liacs.leidenuniv.nl} \hspace{0.1cm} \texttt{zxf@cqut.edu.cn} \hspace{0.1cm} \texttt{cha01nbox@gmail.com}\\
}

\begin{document}
\maketitle

\begin{abstract}
Empathetic response generation aims to comprehend the cognitive and emotional states in dialogue utterances and generate proper responses.
Psychological theories posit that comprehending emotional and cognitive states necessitates iteratively capturing and understanding associated words across dialogue utterances.
However, existing approaches regard dialogue utterances as either a long sequence or independent utterances for comprehension, which are prone to overlook the associated words between them.
To address this issue, we propose an Iterative Associative Memory Model (IAMM)\footnote{Our code is available at \url{https://github.com/zhouzhouyang520/IAMM}} for empathetic response generation.
Specifically, we employ a novel second-order interaction attention mechanism to iteratively capture vital associated words between dialogue utterances and situations, dialogue history, and a memory module (for storing associated words), thereby accurately and nuancedly comprehending the utterances.
We conduct experiments on the Empathetic-Dialogue dataset.
Both automatic and human evaluations validate the efficacy of the model. 
Variant experiments on LLMs also demonstrate that attending to associated words improves empathetic comprehension and expression.
\end{abstract}

\section{Introduction}
As an important task for improving dialogue quality, empathetic response generation aims to comprehend the emotional and cognitive states of the user in dialogue utterances and provide appropriate responses
~\cite{rashkin2018towards,zhong2020towards,liang2021infusing,zheng-etal-2021-comae,liu2021towards}.

The majority of methods treat the dialogue utterances as a long sequence to comprehend the user states ~\cite{lin2019moel,majumder2020mime,CEM2021,li-etal-2022-kemp,zhou2022case}.
These approaches ignore the discrepancies in meanings among individual utterances, leading to inaccurate understanding of emotional and cognitive states~\cite{welivita2021large,wang2022empathetic-seek}.
To address this issue, some methods comprehend more delicate emotional and cognitive states within a set of independent utterances by distinguishing self-other awareness~\cite{zhao2022EmpSOA} or emphasizing emotion-intent transitions~\cite{wang2022empathetic-seek}.

\begin{figure}
\centering
\includegraphics[width=78mm]{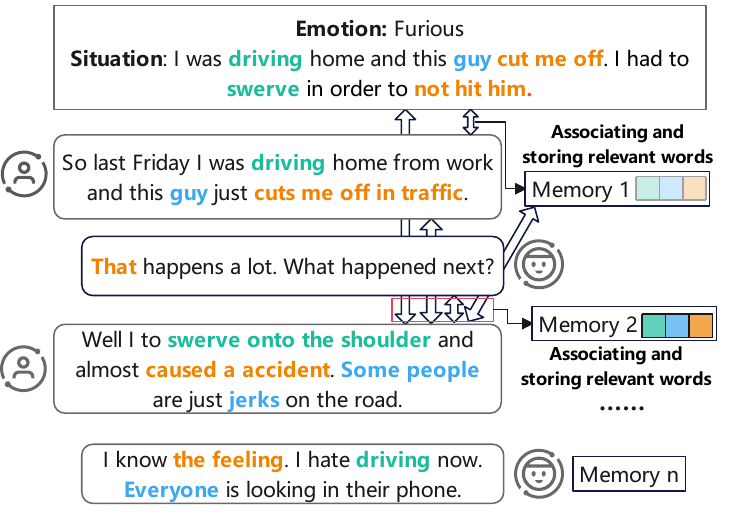}
\caption{\label{fig 1}
An example of iterative association.
Words with the same color are associated. The memory stores the associated words.
}
\end{figure}

However, the situation model~\cite{van1983strategies}, as an important theory for understanding empathy, posits that comprehending emotional and cognitive states in detail necessitates not only understanding independent utterances, but also iteratively associating pivotal associated words within those utterances~\cite{zwaan1998situation,mcnamara2009toward,gernsbacher1992readers,gygax2003representation}.
As shown in Figure \ref{fig 1}, the speaker's independent utterances imply the emotion of ``anger,'' yet the overall expression conveys ``furious.''
If the utterances are understood independently, 
the listener is likely to misinterpret the speaker's emotion as ``anger.''
In contrast, the iterative association integrates subtle associated words, allowing for an accurate understanding of the dialogue.
Specifically, when faced with the first utterance, the listener combines this utterance with related words from the situation and stores it in its memory to form an initial understanding.
When encountering the speaker's second utterance, the listener meticulously compares and reasons this utterance with the dialogue history, situation, and related words in memory, to deepen its understanding of the utterance.
For instance, associating ``jerks'' and ``guy'' reveals an intensification of the emotion of anger, i.e., ``furious''.
Additionally, reasoning that ``cut me off'' and ``caused an accident'' makes it easier to realize the speaker's furious due to the life-threatening event.
Overall, by associating explicit and implicit information, the listener attains a more nuanced understanding of the utterances.
While this comprehension process proves effective, simulating it to achieve meticulous understanding of dialogues remains an open challenge.


In this paper, we propose an iterative associative memory model (IAMM) that iteratively employs an information association module to identify and learn subtle connections within both explicit and implicit information.
We first treat the dialogue content, including both the dialogue utterances and situations, as explicit information, and treat the reasoning knowledge about the dialogue content generated by COMET~\cite{hwang2021comet} as implicit information.
Subsequently, we iteratively utilize the information association module to identify and learn associated words between utterances and situations, dialogue history, and memory (initialized as an empty set) in the explicit/implicit information, and store them in the memory for a thorough understanding of the utterances.
Specifically, the information association module, inspired by the idea that "pages (nodes) linked by important pages (nodes) are also more important"~\cite{brin1998anatomy,weng2010twitterrank}, effectively identifies associated words in the to-be-associated sentences through a second-order interaction attention mechanism.

To validate our model, we construct IAMM and IAMM$_{large}$ (LLMs-based model). 
Experiments are conducted on the Empathetic-Dialogue dataset~\cite{rashkin2018towards}.
Both automatic evaluation and human evaluation demonstrate that compared with the state-of-the-art baselines, our models possess stronger understanding while expressing more informative responses.


Overall, our contributions are as follows:

\begin{itemize}
\item 
We introduce an iterative association framework for empathetic response generation, which simulates the human iterative process of understanding emotions and cognition.

\item 
We propose an iterative associative memory model (IAMM), which iteratively employs a second-order interaction attention mechanism to capture subtle associations in dialogues.

\item 
Experiments on the Empathetic-Dialogue dataset validate the efficacy of our models. 
\end{itemize}

\section{Related Work}
Empathetic response generation requires comprehending the user states in dialogue utterances to generate appropriate responses~\cite{rashkin2018towards}.
Existing methods can be categorized into dialogue-level models and utterance-level models, according to whether they understand dialogue utterances independently.

\textbf{Dialogue-level models}.
Dialogue-level models view all dialogue utterances holistically as a long sequence to comprehend user states.
Some dialogue-level models focus on the coarse emotions~\cite{majumder2020mime,lin2019moel,rashkin2018towards} or subtle emotions~\cite{li2019empdg,li-etal-2022-kemp,gao2021improving,Kim2021empathy,yang2023exploiting} present in a conversation to understand the user states.
While focusing on emotional states, these models ignore cognitive states, leading to inadequate empathy comprehension~\cite{CEM2021}.
Some methods introduce reasoning knowledge to more comprehensively attend to user states~\cite{CEM2021,zhou2022case,cai2023improving}.

\textbf{Utterance-level models}.
Utterance-level models focus on differences in emotional and cognitive states within individual utterances. These models view dialogue utterances as independent sentences to understand the user's state within them via self-other awareness differentiating ~\citet{zhao2022EmpSOA} and emotion-intent transitions ~\citet{wang2022empathetic-seek}.

The aforementioned methods view the dialogue utterances as either long sequences or as independent sentences to understand the user's emotions and cognitive states. 
However, understanding emotional and cognitive states requires iteratively capturing and comprehending associated words~\cite{mcnamara2009toward,gernsbacher1992readers,gygax2003representation}.
Therefore, we propose an Iterative Associative Memory Model (IAMM), which iteratively understands the associated words between utterances from both explicit and implicit information, and generates empathetic responses.

\begin{figure*}
\centering
\includegraphics[width=160mm]{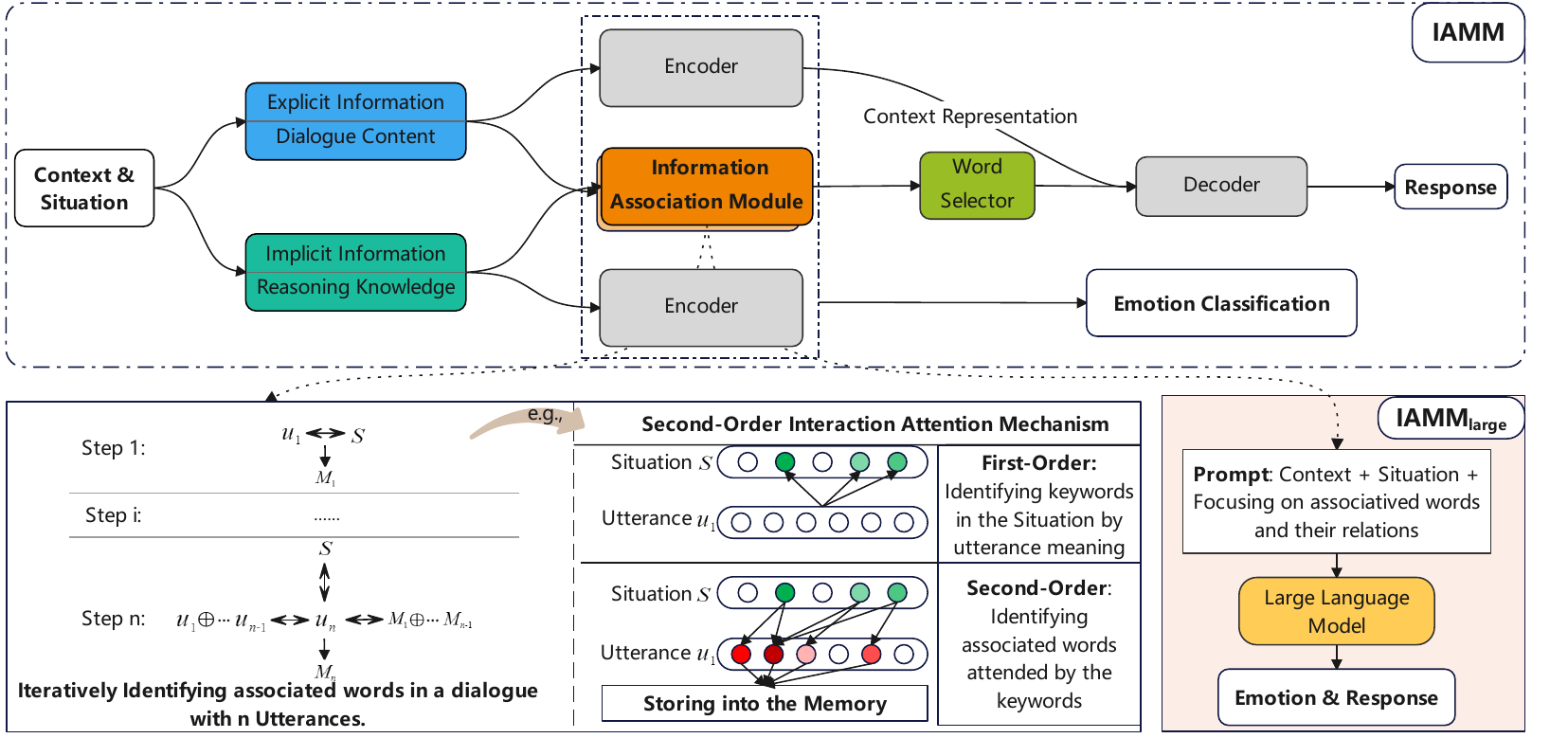}
\caption{\label{fig IAMM}
Overview of IAMM and IAMM$_{large}$, which are small-scale and large-scale models focusing on association information, respectively.
IAMM mainly consists of the following steps: (1) Encoding dialogue information, including explicit information (dialogue utterances and situation) and implicit information (reasoning knowledge); (2) Iteratively capturing associative information, namely associated words between sentences; (3) Predicting emotion and generating responses.
Moreover, IAMM$_{large}$ focuses on subtle associations by injecting associated words into instructions.
}
\end{figure*}
\section{Method}
\subsection{Task Formulation}
The task of empathetic response generation is as follows:
Given the dialogue content, including the context $D$ = [$U_1$, ..., $U_i$, ..., $U_{M}$] and the situation $S$ = [$w^s_1$, $w^s_2$, ..., $w^s_m$], the model needs to understand the dialogue emotion $E$ and generate an appropriate response $Y$ = [$y_1$, $y_2$, ...,$y_j$, $y_N$].
$U_i$ = [$w^i_1, w^i_2, ..., w^i_{m_i}$] is the $i$-th utterance containing $m_i$ words.
$S$ represents a situation description consisting of $m$ words. 
$Y$ represents a response sequence consisting of $N$ words.

\subsection{Overview}
We illustrate the overall architecture of IAMM in Figure \ref{fig IAMM}.
As iterative association requires attention to both explicit and implicit information~\cite{garrod2012referential}, we view the dialogue content as explicit information, and their reasoning knowledge~\cite{hwang2021comet} as implicit information. 
We construct IAMM in the following three steps:
(1) Encoding dialogue information (\textbf{Section \ref{Encoding Dialogue Information}}), which learns explicit and implicit information in dialogue through multiple encoders.
(2) Capturing associative information (\textbf{Section \ref{Capturing Associated Information}}), which iteratively captures associated words in the dialogue regarding the explicit and implicit aspects.
(3) Predicting emotion and generating responses (\textbf{Section \ref{Predicting Emotion and Generating Response}}), which predicts the dialogue emotion and generates responses based on the learned content.
Furthermore, we construct a variant model IAMM$_{large}$ (\textbf{Section \ref{IAMM on Large Language Models}}), which leverages instructions to guide the large language models to focus on associated words and capture the nuanced associations within the dialogue.
\subsection{Encoding Dialogue Information} 
\label{Encoding Dialogue Information}
We encode the dialogue information from both explicit and implicit perspectives.

\textbf{Explicit Information}. We take the dialogue utterances and situations as explicit information.
Following previous work~\cite{wang2022empathetic-seek}, we add special start tokens [CLS] before the dialogue utterances and situations to obtain word sequences $U_i$ and $S$, respectively.
Based on the word sequence $U_i$ of the dialogue utterances, we first learn sentence representations using an utterance-level encoder $Enc_u$, then learn the overall meaning of the dialogue using a dialogue-level encoder $Enc_{ctx}$.
\begin{gather}
H^i_{u} = Enc_{u}(E^i_w + E_p + E_r) \\
H_{c} = Enc_{ctx}(H^1_{u} \oplus ...\oplus H^M_u)
\end{gather}
where $E^i_w$, $E_p$, $E_r$ are the word embedding, positional embedding, and state embedding of the utterances, respectively. The state embedding is used to distinguish between speakers and listeners. 
And $H^i_{u} \in R^{m_i \times d}$, $H_{c} \in R^{M_D \times d}$, with $m_i$, $H_{c}$ being the lengths of the $i$-th utterance and total dialogue utterances.
$d$ and $\oplus$ represent the hidden size and concatenation operation, respectively.

For the word sequence $S$ of the situations, we employ an encoder $Enc_{sit}$ to learn the situation representations.
\begin{gather}
H_{s} = Enc_{sit}(E^{sit}_w + E^{sit}_p)
\end{gather}
where $E^{sit}_w$ and $E^{sit}_p$ represent the word embedding and the positional embedding of the situations, respectively. $H_{s} \in R^{m \times d}$ and $m$ is the number of words it contained.

\textbf{Implicit Information}.
Following \cite{CEM2021, wang2022empathetic-seek}, we utilize the COMET model~\cite{hwang2021comet} to generate reasoning knowledge $K_{u_i}$ and $K_s$ for the dialog utterance $U_i$ and situation $S$, and take them as the implicit information.
Among them, the reasoning knowledge $K_{rk}$ ($rk \in (U_i, s)$) contains 5 types of text:
the impact of events (xEffect), personal emotional reactions (xReact), personal intentions (xIntent), personal needs (xNeed), and personal desires (xWant). For convenience, we represent them as $K^{type}_{rk}$, where $type \in [xEffect, xReact, xIntent, xNeed, xWant]$.

Subsequently, we prepend [CLS] before the reasoning knowledge and feed it into the encoder $Enc_{cs}$.
To understand the overall implicit information, we utilize the encoder $Enc_{cs}$ to learn the reasoning knowledge of the situation and the last utterance.
\begin{gather}
H^{type}_{rk} = Enc_{cs}([CLS] \oplus K^{type}_{rk}) \\
H^{cs}_{rk} = \underset{type}{\Vert} H^{type}_{rk}[0]\\
H_{cs} = H^{cs}_{s} \oplus H^{cs}_{U_M} \\
\widetilde{H}_{cs} = \widetilde{Enc}_{cs}(H_{cs})
\end{gather}
Where $H^{type_i}_{rk}[0]$ is the overall semantic representation of the sentence,
and $rk \in (U_i, s)$, $type \in [xEffect, xReact, xIntent, xNeed, xWant]$.
$\Vert$ is the concatenation operation.
$U_M$ and $s$ represent the last utterance and situation, respectively.
$H^{type}_{rk} \in R^{d^{type}_{rk} \times d}$, $H_{cs} \in R^{10 \times d}$, $\widetilde{H}_{cs} \in R^{10 \times d}$ and $d^{type}_{rk}$ is the number of words in the reasoning knowledge $K^{type}_{rk}$.

\subsection{Capturing Associated Information}
\label{Capturing Associated Information}
We employ the information association module (IAM) to capture important associated words in explicit and implicit information to coherently and thoroughly understand the dialogue.
To elaborate on the process of the information association module, we explain it in two aspects:
Iterative association process, which describes the process executed by the information association module.
Information association module, which explicates in detail the internal structure of the module.

\subsubsection{Iterative Association Process}
To understand the current utterance $U_i$, we construct pairs of explicit and implicit information as the input to the information association module.
These pairs are used to capture important associations between utterance-situation, utterance-dialogue history, and utterance-memory, respectively. Here, $M_i$ and $\overline{M}_i$ represent the memory, which is used to store associated words and is initialized as empty.
Note that when $U_i$=$U_1$, both the dialogue history and memory are empty.

\textbf{Explicit Information Pairs}: 
$(E^i_w,E_s)$, $(E^i_w, E^1_w \oplus ..., E^{i-1}_w)$, $(E^i_w, M_1 \oplus ..., M_{i-1})$.

\textbf{Implicit information pairs}: 
$(H^{cs}_{U_i}, H^{cs}_{s})$, $(H^{cs}_{U_i}, H^{cs}_{U_1} \oplus ..., H^{cs}_{U_{i-1}})$, $(H^{cs}_{U_i}, \overline{M}_1 \oplus ..., \overline{M}_{i-1})$.

Next, we feed the pairs of explicit and implicit information into the information association module (IAM). 
Since the processing procedures for the two types of information are the same, we take the processing procedure for explicit information as an example to illustrate.

Based on the utterance pairs of $U_i$, including utterance-situation, utterance-dialogue history, and utterance-memory, the IAM bidirectionally identifies key associated words in the pairs' sentences and stores them in the memory. 
Subsequently, the model continuously capture the associated words of the next utterance $U_{i+1}$ until the last dialogue utterance.
This iterative process facilitates thoroughly capturing the intricate associations between the utterance and related sentences, thereby enabling better capture of critical information.
\subsubsection{Information Association Module}
This module identifies associated words between relevant sentences, such as the situation and utterance. Inspired by the notion that "nodes highly attended by key nodes are likely important", we propose a \textbf{second-order interaction attention mechanism} with two processes: 
the first-order interaction attention identifies situation words attended by the overall utterance meaning as keywords (key nodes).
The second-order interaction attention identifies utterance words attended by the keywords as associated words (the nodes highly attended by key nodes). 
This two-stage filtering selects associated words that explicitly reflect the connections between the two sentences.

Since associated words exist simultaneously in two sentences (e.g. in the utterance or situation), we bidirectionally select words from the sentences.
For example, selecting associated words in the situation based on the utterance (i.e., u2s), and selecting associated words in the utterance based on the situation (i.e., s2u).
For simplicity, we elaborate on the whole process using the example of s2u:
(1) Constructing association matrices that contain association scores between words and are used for word identification.
(2) First-order Interaction Attention, which identifies keywords in the situation.
(3) Second-order Interaction Attention, which identifies associated words in the utterance based on the keywords in the situation.
(4) Storing the bidirectionally associated words in memory for the next processing.

\textbf{Constructing Association Matrices}. 
To capture rich features, we construct two multi-head association matrices.
$A^{s2t}$ represents the association scores from the situation words to the utterance words, while $A^{t2s}$ represents the association scores from the utterance words to the situation words.
\begin{gather}
A^{s2t}_{ij} = \underset{n=1}{\overset{H}{\Vert}} \sigma((w^n_q v^n_{s,i})^T(w^n_k v^n_{t, j}))\\
A^{t2s}_{ij} = \underset{n=1}{\overset{H}{\Vert}} \sigma((w^n_q v^n_{t, i})^T(w^n_k v^n_{s,j})) 
\end{gather}
where $A^{s2t} \in R^{H \times d_s \times d_t}$, $A^{t2s} \in R^{H \times d_t \times d_s}$ and H is the number of multi-heads.
Here, $\sigma$ is the Sigmoid function.

\textbf{First-order Interaction Attention}.
We identify keywords in the situation.
Intuitively, the words that are associated as important by most other words are more likely to be important. Therefore, we select the words in the situation that have higher degrees of association with utterance words as the keywords.
\begin{gather}
A^s = Mean(A^{t2s})  \\
A^s_{k_1} = \widetilde{Top}_{k_1}(A^s)
\end{gather}
where $Mean$ and $\widetilde{Top}$ are the average function and filter function respectively.
$k_1$ represents the number of important words to filter.
And $A^s \in R^{H \times d_s}$, $A^s_{k_1} \in R^{H \times k_1}$.

\textbf{Second-order Interaction Attention}.
We select words in the utterance that have the highest degrees of association with the key situation words, as important associated words.
\begin{gather}
A^{s2t}_{k_2}, E^{s2t}_{k_2} = Top_{k_2}(A^{t2s}) \\
\widetilde{E}^{s2t}_{k_2} = \underset{k=1}{\overset{k_2}{\Vert}} A^{s2t}_k \cdot E^{s2t}_k \\
E^{st}_t = A^s_{k_1} \cdot \widetilde{E}^{s2t}_{k_1,k_2}
\end{gather}
where $A^{s2t}_{k_2} \in R^{H \times d_s \times k_2}$, $E^{s2t}_{k_2} \in R^{H \times d_s \times k_2 \times d_h}$, 
and they are the representations and scores of the utterance words attended by situation words, respectively.
$K2$ is the number of associated words.
$\widetilde{E}^{s2t}_{k_2} \in R^{H \times d_s \times (k_2 \times d_h)}$, $E^{st}_t \in R^{H \times k_1 \times (k_2 \times d_h)}$ and $d_h$ is the hidden size.
$E^{st}_t$ is the representation of associated words.

We select the associated words from the situation in the same way. Subsequently, we concatenate the associated words of the situation and the utterance, where $E_{st} \in R^{(2H \times k_1) \times (k_2 \times d_h)}$.
\begin{gather}
E_{st} = E^{st}_t \oplus E^{st}_s
\end{gather}

\textbf{Storing the Associated Words in Memory}.
Regarding explicit information, we select associated words between utterances and situations ($E^{ek}_{sc}$), between utterances and dialogue history ($E^{ek}_{cc}$), and between utterances and memory ($E^{ek}_{mc}$) in the same way.
\begin{gather}
V_{ek} = E^{ek}_{sc} \oplus E^{ek}_{cc} \oplus E^{ek}_{mc}
\end{gather}
where $E^{ek}_{sc}, E^{ek}_{cc}, E^{ek}_{mc} \in R^{(2H \times k_1) \times (k_2 \times d_h)}$.

Specifically, regarding implicit information, we also construct associated words between utterances and situations ($E^{ik}_{sc}$), between utterances and dialogue history ($E^{ik}_{cc}$), and between utterances and memory ($E^{ik}_{mc}$). 
When iterating to the last sentence, we combine the explicit information memory $V_{ek}$ and the implicit information memory $V_{ik}$ as the final associated information $V$.
\begin{gather}
V_{ik} = E^{ik}_{sc} \oplus E^{ik}_{cc} \oplus E^{ik}_{mc} \\
V = V_{ek} \oplus V_{ik}
\end{gather}
where $E^{ik}_{sc}, E^{ik}_{cc}, E^{ik}_{mc} \in R^{(2H \times k_1) \times (k_2 \times d_h)}$.
$V \in R^{L_m \times d}$ and $L_m$ is the number of associated words in the memory.

Based on the memory of associated words, we learn association information through an encoder $Enc_m$.
\begin{gather}
H_m = Enc_{m}(V)
\end{gather}

\subsection{Predicting Emotion and Generating Response}
\label{Predicting Emotion and Generating Response}
\subsubsection{Prediction Emotion}
In order to predict emotion, we input the dialogue utterance representation $H_c$, the situational representation $H_s$, 
into aggregation network $AN_u$ \cite{yang2023exploiting} to obtain emotion representations. We subsequently use these representations to separately predict the probability of emotions.
\begin{gather}
P_{c} = \phi(AN_{u}(H_{c})) \\
P_{s} = \phi(AN_{u}(H_{s}))
\end{gather}
where $\phi$ is the softmax function.
$P_{c}, P_{s} \in R^{d_e}$ and $d_e$ is the number of emotions.

Similarly, we also predict the emotion probabilities for the association representation $H_m$, and the reasoning knowledge representation $H_{cs}$.
\begin{gather}
P_{a} = \phi(AN_{a}(H_m)) \\
P_{cs} = \phi(AN_{cs}(\widetilde{H}_{cs}))
\end{gather}
where $P_{a}, P_{cs} \in R^{d_e}$.
$AN_a$, and $AN_{cs}$ represent aggregate networks with the same architecture but different parameters.

We multiply the above emotion probabilities as the final emotion probability. Then we use log-likelihood loss to optimize the parameters based on the emotion probability and the ground truth label $e^*$.
\begin{gather}
    P_e = P_c(e^*) \cdot P_s(e^*) \cdot P_a(e^*) \cdot P_{cs}(e^*) \\
    \mathcal{L}_e = - log(P_e)
\end{gather}
\subsubsection{Generation Response}
To fully utilize the associative words containing important information, we design a word selector and flexibly incorporate associative information during decoding.

\textbf{Word Selector}. We select important associated words to utilize the effective information in memory.
\begin{gather}
    S_m = \sigma(w_v H_m) \\
    \widetilde{S}_m, \widetilde{H}_m = Top_{k_3}(S_m, H_m) \\
    \widetilde{H}_m = \widetilde{S}_m \cdot \widetilde{H}_m
\end{gather}
where $Top$ is a selecting function that selects the top associated words with the highest scores from the memory $H_m$ based on the score $S_s$.
$w_v \in R^{d \times 1}$ is a trainable parameter.
$\widetilde{H}_m \in R^{k_3 \times d}$ and $K_3$ is a hyperparameter.

Based on the representations of the utterance and associated words, we generate the decoding vectors $O_{c}$ and $O_{s}$ , respectively.
\begin{gather}
    O_{c} = Dec_{c}(H_c) \\
    O_{m} = Dec_{a}(\widetilde{H}_m)
\end{gather}
where $Dec_c$ and $Dec_a$ represent decoders with the same architecture but different parameters.
$O_c$, $O_m$ $\in R^{L_t \times d}$, and $L_t$ is the length of the response words at time $t$.

We then combine the decoding vectors to form $O$ and use it to predict word probabilities.
\begin{gather}
    g = \sigma(w (O_c \oplus O_m)) \\
    O = g \cdot O_c + (1 - g) \cdot O_{m} \\
        P(y_t|y<t, D, S) = Generator(E_{y<t}, O)
\end{gather}
where $w \in R^{d \times 1}$ represents learnable parameters. 
$Generator$ denotes the point generator \cite{see2017getPointGenerator} that transforms the decoded vectors into word probabilities.

Finally, we employ cross-entropy loss as the generation loss $\mathcal{L}_{gen}(y_t)$. By integrating the generation loss $\mathcal{L}_{gen}(y_t)$ and the emotion loss $\mathcal{L}_e$, we optimize the overall parameters.
\begin{gather}
    \mathcal{L}_{gen}(y_t) = - \sum^T_{t=1}log(P(y_t|y<t, D, S)) \\
    \mathcal{L} = \mathcal{L}_{gen}(y_t) + \mathcal{L}_e
\end{gather}


\subsection{Baselines}
To compare the performance of IAMM, we select the state-of-the-art models as baselines. 
\textbf{EmpDG}~\cite{li2019empdg} considers fine-grained emotional words and user feedback;
\textbf{KEMP}~\cite{li-etal-2022-kemp} enhances hidden emotional representations using a ConceptNet-based emotional graph;
\textbf{CEM}~\cite{CEM2021} enhances emotion and cognition using reasoning knowledge;
\textbf{CASE}~\cite{zhou2022case} aligns emotion and cognition from fine-grained and coarse-grained perspectives;
\textbf{SEEK}~\cite{wang2022empathetic-seek} is a model that captures emotional-intention transitions in dialogue utterances;
\textbf{ESCM}~\cite{yang2023exploiting} captures emotion-semantic dynamic associations based on word-level emotions.

\subsection{Implementation Details}
We conduct experiments on the EMPATHETIC-DIALOGUES~\cite{rashkin2018towards} dataset.
The details are provided in Appendix \ref{appendix A}.

\subsection{Evaluation Metrics}
To comprehensively understand model performance, we conduct automatic and human evaluations.  

\textbf{Automatic Evaluation}. 
Following previous methods~\cite{li-etal-2022-kemp, CEM2021}, we use perplexity (PPL), accuracy (Acc), distinct-1 (Dist-1)/distinct-2 (Dist-2)~\cite{li2015diversity} to evaluate response fluency, emotion classification accuracy, and response diversity, respectively. Specifically, lower perplexity indicates better quality, while higher values are better for the other metrics.

\begin{table}
\centering
\begin{tabular}{ccccc}
\hline
\textbf{Models} & \textbf{Acc} & \textbf{PPL} & \textbf{Dist-1} & \textbf{Dist-2} \\
\hline
EmpDG & 34.31 & 37.29 & 0.46 & 2.02 \\ 
KEMP & 39.31 & 36.89 & 0.55 & 2.29 \\ 
CEM & 39.11 & 36.11 & 0.66 & 2.99 \\ 
CASE & 40.2 & 35.37 & 0.74 & 4.01 \\ 
SEEK & 41.85 & 37.09 & 0.73 & 3.23 \\
ESCM & 41.19 & \textbf{34.82} & 1.19 & 4.11 \\
\hline
IAMM & \textbf{55.92} & 35.66 & \textbf{2.09} & \textbf{7.03} \\ 
\hline
w/o EA
& 52.43 & 35.17 & 1.18 & 4.04\\

w/o IA
& 51.48 & 35.29 & 1.54 & 5.14\\

w/o WS
& 55.82 & 35.46 & 1.75 &  6.01\\
\hline
\end{tabular}
\caption{\label{table automatic}
Results of automatic evaluation.
}
\end{table}

\begin{table}
\centering
\begin{tabular}{ccccc}
\hline
\textbf{Comparisons} & \textbf{Aspects} & \textbf{Win} & \textbf{Lose} & \textbf{$\kappa$} \\
\hline

\hline
\multirow{3}{*}{\centering \shortstack{IAMM \\ vs. ESCM}} & Emp. & \textbf{23.5} & 20.0 & 0.43\\
& Rel. & \textbf{24.4} & 22.7 & 0.42 \\
& Flu. & \textbf{18.0} & 16.8 & 0.43 \\
\hline
\multirow{3}{*}{\centering \shortstack{IAMM \\ vs. SEEK}} & Emp. & \textbf{33.6} & 20.2 & 0.46 \\
& Rel. & \textbf{41.2} & 16.8 & 0.45 \\
& Flu. & \textbf{18.5} & 16.5 & 0.43 \\
\hline
\end{tabular}
\caption{\label{table human}
Results of human evaluation, where $\kappa$ is the inter-labeler agreement measured by Fleiss's kappa~\cite{fleiss1973equivalence}, and 0.4 $\textless$ $\kappa$ $\leq$ 0.6 indicates moderate agreement.
}
\end{table}
\textbf{Human Evaluation Metrics}.
We invite three professional crowdworkers to evaluate the response quality.  
Consistent with previous methods~\cite{majumder2020mime, yang2023exploiting}, we conduct A/B testing to compare the baseline and IAMM.
For a response, if IAMM has better quality, the crowdworkers add one point to \emph{Win}. 
If IAMM is worse than the comparison model, they add one point to \emph{Lose}.
To evaluate quality, we consider empathy (Emp.), relevance (Rel.), and fluency (Flu.):
Empathy measures whether the emotion in the response is appropriate.
Relevance measures whether the response is relevant to the dialogue topic and content.
Fluency measures whether the language of the response is natural and fluent.

\section{Results and Analysis}
\subsection{Main Results}





\textbf{Automatic Evaluation Results}.
As shown in Table \ref{table automatic}, IAMM outperforms the baselines on most metrics.
For emotion accuracy, IAMM significantly surpasses the baselines.
This is because leveraging associated information facilitates dialogue subtle understanding, which promotes accurate emotional comprehension.
In diversity, IAMM substantially surpasses the baselines.
This is primarily attributed to the focused important associated words facilitating the generation of informative responses.
Regarding perplexity, IAMM does not surpass the baseline. This is because the generated responses contain fewer generic sentences could have better perplexity and lower diversity. For instance, for the context ``I am very ashamed in my grades'', responses like ``I am so sorry to hear that'' may have better PPL and lower diversity, but they are not relevant or empathetic to the context. In contrast, responses like ``I know that feeling. I have a bad grade and I know how you feel'' may have worse PPL compared to the former, but it is more likely to be relevant and empathetic to the context.

\textbf{Human Evaluation Results}.
As shown in Table \ref{table human}, IAMM also demonstrates better performance over the baselines in human evaluation.  
The model shows superiority in empathy.
This is because focusing on the associations between utterances promotes delicate understanding, thus expressing appropriate emotional responses.
The model is excellent in relevance.
This is primarily because associated words tend to be important words, and responses generated based on these significant words are more likely to be relevant.
Regarding fluency, IAMM is better than the baselines in human evaluation, although it is worse on the automatic metric (PPL).  
This is primarily because the automatic metric (PPL) tends to favor sentences that are common in the training set, while human evaluation conforms better to natural language.
IAMM generates fewer common and general sentences, such as "I am so sorry to hear that." Hence, the evaluation results turn out this way.

\subsection{Ablation Studies}
As shown in Table \ref{table automatic}, we conduct the following ablation experiments to validate the effectiveness of each module:
\textbf{w/o EA}: without explicit association;
\textbf{w/o IA}: without implicit association;
\textbf{w/o WSD}: without the associated word selector and decoder.

For emotion accuracy and diversity, the results show both explicit and implicit associative information have considerable influence.
Explicit associative information contributes more to diversity, while implicit information contributes more to emotion inference.
This indicates that associated words in explicit information are more easily utilized for expression, while those in implicit information are more conducive for emotion inference.
For perplexity, incorporating both types of information leads to better perplexity, as more non-general informative expressions are generated.
Furthermore, without the associated word selector and decoder, the model's diversity decreases, owing to insufficient expression of key information.

\subsection{IAMM on Large Language Models}
\label{IAMM on Large Language Models}
\begin{table}
\centering
\begin{tabular}{cccc}
\hline
\textbf{Models} & \textbf{Acc} & \textbf{Dist-1} & \textbf{Dist-2} \\
\hline
GLM3$_{7B}$ & 62.16 & 3.51 & 21.56 \\
IAMM$_{GLM}$ & \textbf{62.6} & \textbf{3.55} & \textbf{21.7} \\ 
\hline
GPT3.5 & 37.9 & 3.58 & 21.38 \\
IAMM$_{GPT}$ & \textbf{38.51} & \textbf{3.63} & \textbf{22.13} \\
\hline
\end{tabular}
\caption{\label{table llms}
Automatic evaluation results of IAMM on large language models.
}
\end{table}
Large language models have shown superior performance on multiple tasks~\cite{qin2023chatgpt,wang2023robustness,chen2023llm,sun2023chatgpt,zheng2023lmsys,tang2023science}.
To further verify the effect of iterative association on large models, we first make the large model pay attention to the associated words captured by IAMM. 
We then use instructions to focus on the relationships between the associated words to deeply understand the dialogue.
The verification methods include fine-tuning and non-fine-tuning.
Regarding the fine-tuning method, we input the associated words through instructions on the Chinese-English mixed model ChatGLM3~\cite{zeng2022glm,du2022glm} and conducted fine-tuning training. 
Regarding the non-fine-tuning method, we instruct GPT-3.5 to pay attention to the associated words and their relationships to enhance the model.

The experimental results are shown in Table \ref{table llms}. Compared with the baseline models, the models focusing on associative relationships have stronger emotion recognition and expression abilities, which further demonstrates the effectiveness of iterative associations.

\subsection{Analysis of Associated Words}
To further explore the characteristics of associated words, we collected 8,012 associated words on the test set and statistically analyzed their emotion intensity and inverse document frequency (IDF) \footnote{Characteristics of IDF: The more common a word is, the lower its IDF score, and vice versa.}.
The analysis results show:
(1) The most frequently attended words are common words (e.g., "that", "guys"). 
(2) Words that are paid more attention to (assigned higher weights) have either greater emotional intensity or are non-stop words (e.g. "traffic", "accident").
The main reasons are:  
(1) The model tends to correlate phrase-phrase and phrase-word associations, such as (swerve onto the shoulder, caused a accident) and (cuts me off in traffic, That) in Example \ref{fig 1}.  
In the phrase-word associations, phrases often contain both common and uncommon words, while the individual words are typically common words.
This results in more common words attended by the model.
(2) The model places more emphasis on phrases with emotion or definite meanings. 
To better understand the emotion or meaning, the model often assigns higher weights to emotion words or uncommon words within the phrase.
See Appendix \ref{appendix C} for details.

\subsection{Case Study}
We conduct case studies for the strongest baseline and IAMM.  
See Appendix \ref{appendix B} for details.

\section{Conclusion and Future Work}
In this paper, we have proposed an Iterative Associative Memory Model (IAMM) for empathetic response generation, inspired by the human iterative process of understanding emotions and cognition.
It employs a novel second-order interaction attention mechanism to iteratively identify key associated words across dialogue utterances, enabling a more accurate understanding of the emotional and cognitive states.
Automatic and human evaluations demonstrate that IAMM accurately understands emotions and expresses more empathetic responses.
Experiments based on large language models and associated word analysis further validate the effectiveness of the iterative associations. In the future, we will explore empathetic comprehension mechanisms based on large language models.



\section{Limitations}
The limitations of our work are as follows: 
(1) This work is inspired by the text-based empathetic comprehension mechanism. As the better comprehension of empathy relies on multimodal and large language models, we will conduct research combining these aspects in the future.
(2) Iterative association relies on situation information. 
Although it is prevalent and effective, some datasets still lack this feature. In the future, we will also explore how to effectively construct situation information.

\section{Ethical Considerations}
Regarding the potential ethical impacts of our work:
(1) The dataset we use is EMPATHETIC-DIALOGUE, which is open source and does not involve any potential ethical risks.
(2) The baseline models we use are also public and do not have potential moral impacts. 
Moreover, the components employed in our model are open-sourced or innovative and do not involve potential ethical risks.

\section*{Acknowledgments}
We are grateful to the reviewers for their diligent evaluation and constructive feedback, which helped enhance the quality of this paper. 
We also appreciate the insightful discussions and comments from the authors, which stimulated valuable thinking and contributed significantly to the development of this research.
This work was supported by National Natural Science Foundation of China (No.61976054).


\bibliography{custom}
\bibliographystyle{acl_natbib}

\appendix
\section{Appendix A}
\label{appendix A}
We conduct experiments on the EMPATHETIC-DIALOGUES~\cite{rashkin2018towards} dataset, which contains 32 types of emotions, i.e. $d_e$ = 32.
In the encoder, the hidden size is d = 300.
In the information association module, the numbers of heads and hidden layer size are $H$ = 2 and $d_h$ = 20, respectively.
The numbers of keywords and associated words are $k_2$ = 15 and $k_1$ = 5, respectively. 
And the number of selected keywords for the decoder is $k_3$ = 5.
The batch size of the model is set to 16.
We optimize the model using Adam optimizer~\cite{kingma2014adam} on an NVIDIA Tesla T4 GPU.  
The model converges after 14,400 iterations.
\section{Appendix B}
\label{appendix B}
We select the strongest baselines to compare with IAMM. The details of the results are listed in Table \ref{table case}.

In the first case, SEEK fails to correctly express the emotion of ``proud''.  
ESCM incorrectly recognizes the emotion within.  
While IAMM perceives the speaker's feeling of ``proud'' by identifying the associative words ``accepted into harvar''. 
At the same time, it also understands the subject stated by the speaker, ``my daughter'', thus incorporates emotions and responses appropriately in the response ``bet she was so proud of her''.

In the second case, as the models fail to identify key information in the dialogue, SEEK and ESCM express general sentences. 
While IAMM discovers the key information ``family'' in the dialogue by associating ``my family'' and ``they''. 
At the same time, by associating the emotion "ashamed" in the sentence, IAMM also clearly understands the emotion.
Based on the key information and emotion, the model generates an empathetic response.

Overall, the iterative associated words in dialogues facilitate nuanced understanding.
Additionally, by conveying key associated words, IAMM produces more informative and relevant responses.

\begin{table*}
\centering
\begin{tabular}{p{1.3cm}|p{12cm}}

\hline
\textbf{Emotion} & \textbf{Proud}\\
\textbf{Situation} & I was so \textcolor{c7}{excited} of \textcolor{green}{my daughter} when \textcolor{green}{she} was \textcolor{c9}{accepted into harvard}. \textcolor{green}{She} worked so hard through high school and has \textcolor{c9}{achieved her dream}.\\
\textbf{Speaker} & \textcolor{green}{My daughter} was \textcolor{c9}{accepted into harvard}. \textcolor{green}{She} starts in the fall.\\
\textbf{Gold} & Any employer would hire someone who went to harvard in a heartbeat. You are a lucky parent!\\
\hline
\textbf{SEEK} & Wow, that is a very long time.\\
\textbf{ESCM} & Oh wow, that is a shame.\\
\hdashline
\textbf{\textbf{\textcolor{blue}{IAMM}}} & That is \textcolor{c7}{awesome}! I bet \textcolor{green}{she} was so \textcolor{c9}{proud of} \textcolor{green}{her}!\\

\hline
\textbf{Emotion} & \textbf{Ashamed}\\
\textbf{Situation} & I am very \textcolor{c9}{ashamed} in \textcolor{green}{my family}.\\
\textbf{Speaker} & I am very \textcolor{c9}{ashamed} in \textcolor{green}{my family}.\\
\textbf{Listener} & I am really \textcolor{c9}{sorry} to hear that. Why are you \textcolor{c9}{ashamed}?\\
\textbf{Speaker} & Because \textcolor{green}{they} never invite me to go places, and \textcolor{green}{they} never call me.\\
\textbf{Gold} & That is really too bad. Did you all have a fight?\\
\hline
\textbf{SEEK} & I am sure you will do well.\\
\textbf{ESCM} & I am sorry to hear that.\\
\hdashline
\textbf{\textbf{\textcolor{blue}{IAMM}}} & I know \textcolor{c9}{that feeling}. I have \textcolor{green}{a family} and I know how you \textcolor{c9}{feel}.\\


\hline
\end{tabular}
\caption{\label{table case}
Case Study of IAMM and Benchmarks, where words with the same color have certain association relationships.
}
\end{table*}

\section{Appendix C}
To validate the characteristics of the associated words, we extracted 8,012 associated words from the test set that the model paid attention to. We statistically analyzed the data in terms of emotions and word frequency.

\begin{figure}
\centering
\includegraphics[width=82mm]{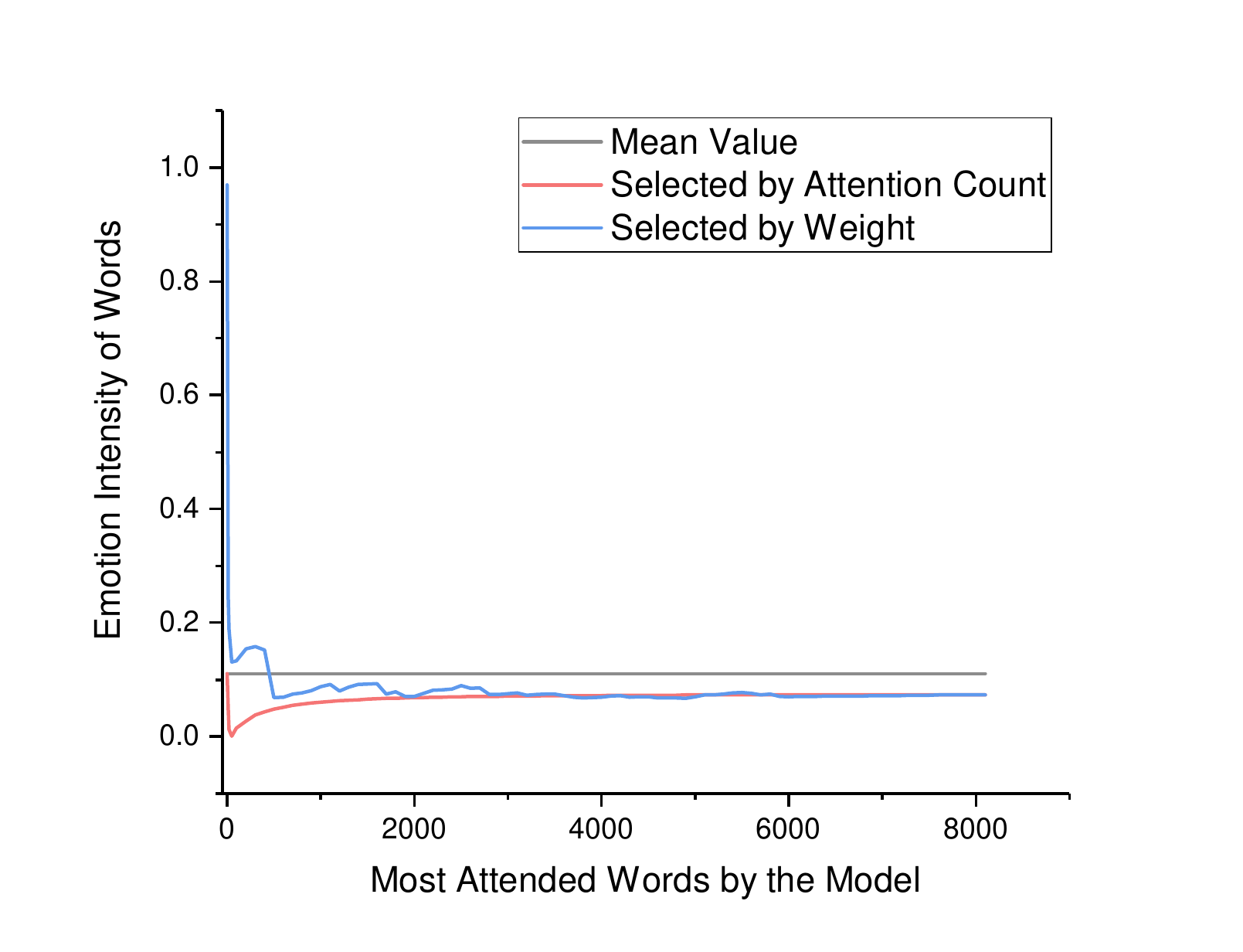}
\caption{\label{fig emo_statistic}
Results of emotion analysis for associated words.
}
\end{figure}

Figure \ref{fig emo_statistic} shows the results of the emotion analysis.
The red line ranks words by attention counts from high to low, and calculates the average emotion intensity of the top $k$ words.
The blue line ranks words by the weight given by the model, and calculates the emotion intensity of the top $k$ words.
The gray line is the average emotion intensity of all words in the test set.
The x-axis is the number of top $k$ words, and the y-axis is the emotion intensity value.
The red line indicates that the most attended words by the model have relatively low emotion intensity, while the blue line indicates that the most attended words by the model have high emotion intensity.

\begin{figure}
\centering
\includegraphics[width=82mm]{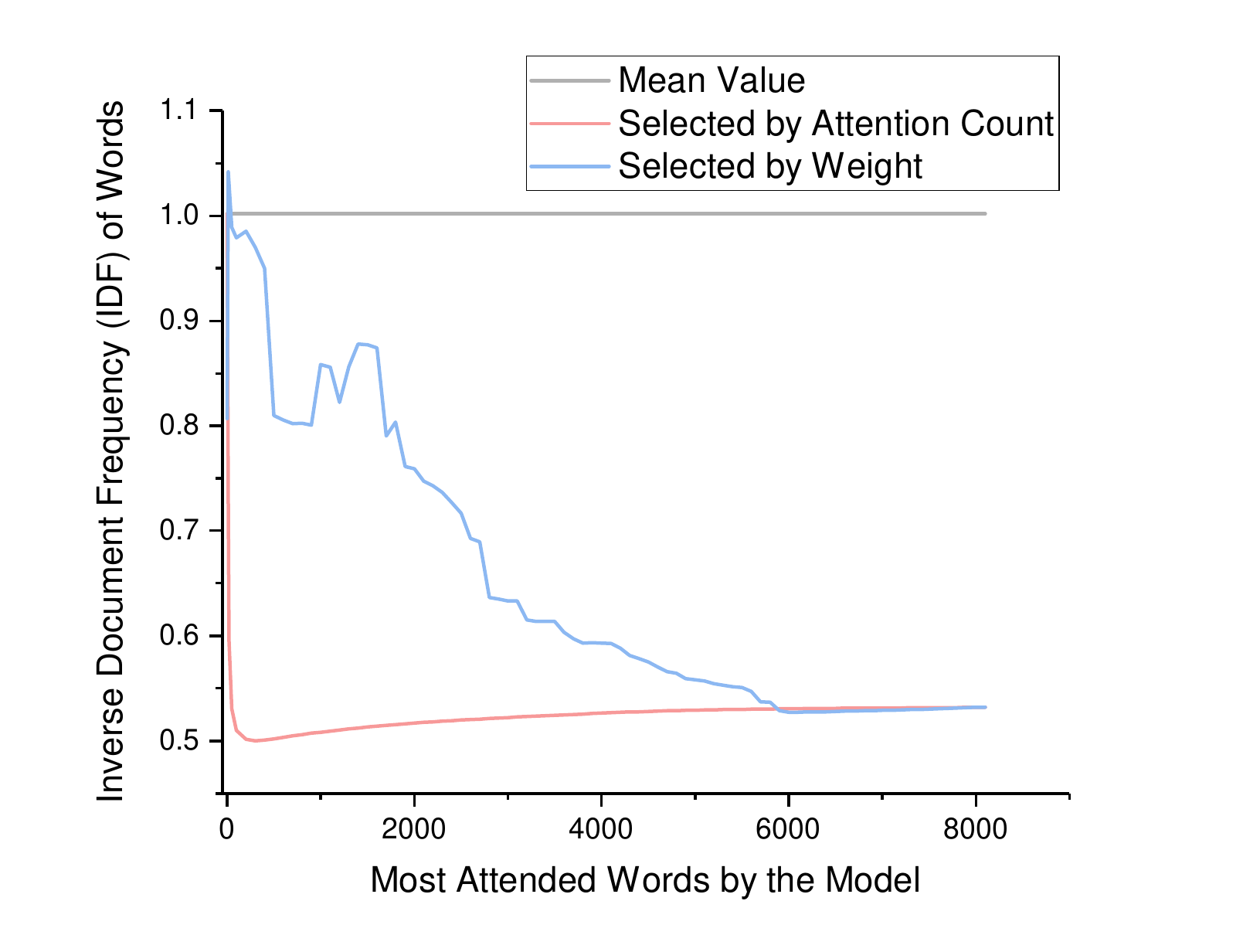}
\caption{\label{fig idf_statistic}
Results of frequency analysis for associated words.
}
\end{figure}

Figure \ref{fig idf_statistic} shows the results of word frequency analysis.
The red line sorts words by attention counts and calculates the average inverse document frequency (IDF) of the top $k$ words.
The blue line sorts words by the weight given by the model, and the gray line is the average IDF for all words.
Similarly, the x-axis represents the number of top $k$ words, and the y-axis represents the IDF value.
The red line indicates that the most attended words are the most common ones, while the blue line indicates that the most attended words are uncommon.

The overall results show that the model pays attention to common words with low emotions (e.g. "That", "it"), while its most highly weighted words have high emotion intensity or are less common.

\label{appendix C}

\end{document}